\documentclass{article}

\usepackage[numbers]{natbib}

\usepackage {arxiv}
\usepackage{cite}
\usepackage{floatrow}
\usepackage{amsmath,amssymb,amsfonts}
\usepackage{caption}
\usepackage{algorithm}
\usepackage{algorithmic}
\usepackage{textcomp}
\usepackage[table]{xcolor}
\usepackage{float}
\usepackage{booktabs}
\usepackage{enumitem}
\usepackage{setspace}
\usepackage[]{graphicx}
\usepackage{amsmath}
\usepackage{amssymb}
\usepackage{epsfig}
\usepackage{epstopdf}

\usepackage{subfigure}
\usepackage{caption}
\usepackage{comment}
\usepackage{multirow}

\title{\LARGE \bf
Between-Sample Relationship in Learning Tabular Data Using Graph and Attention Networks}
\author{Shourav B. Rabbani and Manar D. Samad\\
Department of Computer Science\\
Tennessee State University\\
Nashville, TN, USA\\
\texttt{msamad@tnstate.edu} \\
 }

\begin{document}




\maketitle

\begin{abstract}

 Traditional machine learning assumes samples in tabular data to be independent and identically distributed (i.i.d). This assumption may miss useful information within and between sample relationships in representation learning. This paper relaxes the i.i.d assumption to learn tabular data representations by incorporating between-sample relationships for the first time using graph neural networks (GNN). We investigate our hypothesis using several GNNs and state-of-the-art (SOTA) deep attention models to learn the between-sample relationship on ten tabular data sets by comparing them to traditional machine learning  methods. GNN methods show the best performance on tabular data with large feature-to-sample ratios. Our results reveal that attention-based GNN methods outperform traditional machine learning on five data sets and SOTA deep tabular learning methods on three data sets. Between-sample learning via GNN and deep attention methods yield the best classification accuracy on seven of the ten data sets, suggesting that the i.i.d assumption may not always hold for most tabular data sets.

\end{abstract}

\keywords {embedding clustering, tabular data, Gaussian clusters, autoencoder, representation learning, multivariate distribution}

\section {Introduction}

In a node-edge-like graph structure, edges represent the connection between data samples denoted by nodes. Graph data structures represent individual node samples in terms of their connected neighboring samples in bioinformatics, software vulnerability identification, and communication networks. A graph neural network (GNN) uses neighborhood connectivity and node feature vectors to learn a node representation for node classification~\citep{Zhang2020}. GNN assumes that node samples in a neighborhood are related at varying degrees, subsequently characterizing individual samples. This between-sample relationship is not considered for data samples generally assumed to be independent. For example, tabular data sets are collections of samples in rows with features positioned in columns. In tabular data, individual rows are assumed to be independent and identically distributed (i.i.d.) for machine learning purposes~\citep{ Hamilton2020,Vovk1999}. It is important to note that each tabular data set is unique depending on the application domain. A bioinformatics data set can be structured in a tabular format concealing the relationships between samples. Furthermore, attention learning can focus on how individual samples attend to other samples to find between sample relationship~\citep{Somepalli2022}.

We hypothesize that traditional machine or deep learning methods can disregard useful relationships between samples that can improve the representation of individual samples in certain domains. In other words, GNN and attention-based methods can be competitive for certain tabular data sets over traditional feature-based learning with the i.i.d assumption. This paper investigates whether GNN and attention models can learn hidden relationships between samples in tabular data to improve the baseline classification accuracy obtained using feature vectors alone.

The remainder of the manuscript is organized as follows. Section II provides a theoretical background on the graph and attention networks and a literature review on deep learning methods for tabular data. Section III describes methods for creating an adjacency matrix for learning tabular data on graph networks, the baseline graph and deep learning methods, experimental steps, and evaluation. Section IV summarizes the results by comparing the performance of different graph and attention-based methods on tabular data sets. Section V summarizes the findings, and the conclusions are provided in Section VI.

\section {Background}
\label{background}

\subsection {Graph-based learning}

Let $G = (V, E)$ be a graph where V is the set of $N$ nodes, and $E$ is the set of edges connecting two nodes. The connectivity between nodes is represented by edges and stored in an $N\times N$ adjacency matrix (A) for graph-based computing. The rows and columns of an adjacency matrix refer to the source and destination nodes of a directed graph, respectively. The edges in an undirected graph are numbered by 0 (not connected) and 1 (connected) in a symmetric binary or unweighted adjacency matrix. Edges can measure the association between two nodes using a value greater than zero. A static adjacency matrix is the main input to a GNN for node representation learning. Additionally, individual nodes can be represented by an m-dimensional feature vector. Therefore, Graph Neural Networks (GNNs) can classify nodes by simultaneously learning from feature vectors and the node connectivity information presented in the adjacency matrix. 

\begin{figure*}[t]
\centering
 \includegraphics[trim=0cm 0cm 0cm 0cm, width=1\textwidth] {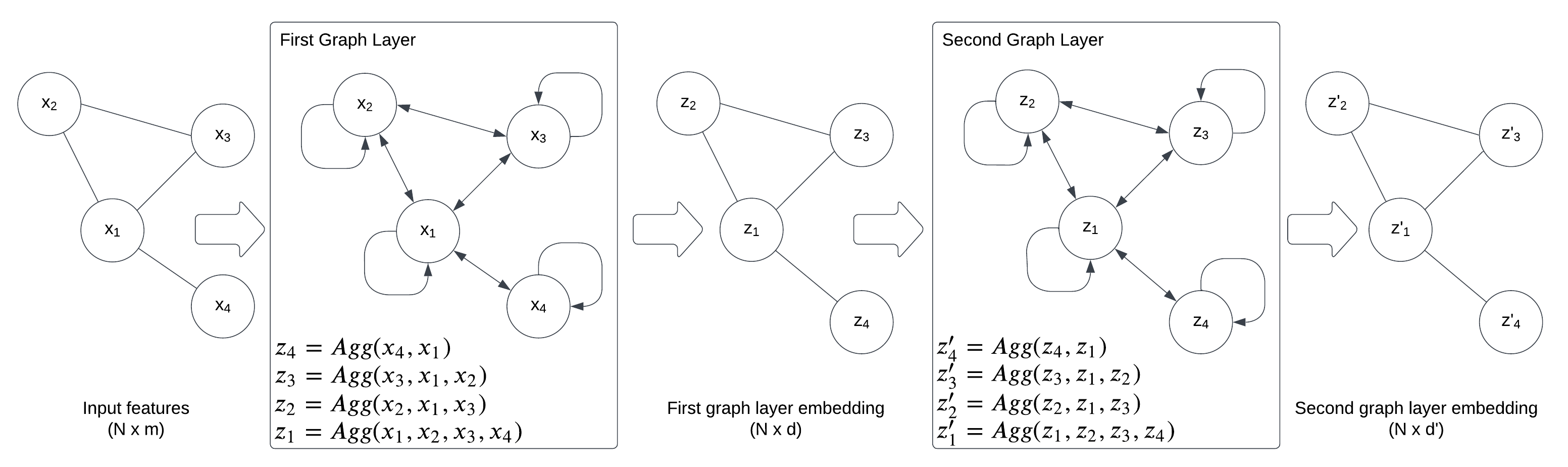}
\caption {A graph layer aggregating neighboring information via message passing. Features of the node itself (self loop) and from neighbors (arrows point towards nodes) are aggregated to create embedding ($z_1$, $z_2$, $z_3$, $z_4$) in a graph layer.}
\label{figure_graph_layer}
\end{figure*}

A GNN model comprises sequential graph layers where each layer aggregates node features from its ‘first-degree’ neighbors, including the node itself, via a message-passing mechanism. The first-degree neighbors of node $x_3$ are nodes $x_1$ and $x_2$, as shown in Figure  \ref{figure_graph_layer}. In the second graph layer, nodes $z_1$, $z_2$, $z_3$, and $z_4$ represent the embeddings corresponding to nodes $x_1$, $x_2$, $x_3$, and $x_4$, respectively. Here, node $z_3$ (embedding of node $x_3$) will now access the information about its second-degree neighbor, $x_4$, via its first-degree neighbor, $z_1$. Therefore, a GNN model with a second layer will encode information from up to the second-degree neighbors. The sequence of graph layers thus encodes an m-dimension input feature vector to a d-dimensional embedding. The final graph layer maps the d-dimensional embedding to the likelihood of $K$ dimensional target labels for a supervised classification task. The embedding update at each graph layer ($l$) is shown in Equation \ref{equation-graph-layer}.
\begin{equation}
H^{l+1}  = \sigma ( A \times H^l \times W^l).
\label{equation-graph-layer}
\end{equation}
Here, $A \in \Re^{N \times N}$ is the adjacency matrix , $H^l \in \Re^{N \times d_l}$, $W^l \in\Re^{d_l \times d_{l+1}}$, $\mathcal{H}^0$ is the input feature matrix where $d_0 = m$ represents the dimensionality of input features. Also, $\sigma$ is the element-wise non-linear activation function, $l$ is the graph layer index where $l = 0,1,2,..., L-1 $), and $L$ is the number of graph layers. The $l$-th graph layer maps $d_l$-dimensional embedding to $d_{l+1}$ dimensional embedding via message passing. To initiate the message passing between nodes, the diagonals of the adjacency matrix are set to $1$ representing self-connections. The presence of many neighbors (high-degree nodes) or large edge weights can inflate the magnitude of the node embedding following the aggregation, which requires neighborhood normalization to avoid numerical instability. Equation \ref{equation-norm-embedding} shows how the embedding of node $p$ as $h_p$ is obtained by aggregating embeddings ($h_q$) of $q$ neighboring nodes. The  neighborhood embeddings are aggregated and normalized by the total number of neighboring nodes ($N_p$). Other normalization methods include symmetric normalization proposed in graph convolutional network (GCN) \citep{Kipf2016}.
\begin{equation}
h_p  = \frac{\sum_{q \in \mathcal{N}_p} h_q}{ \lvert \mathcal{N}_p \rvert }.
\label{equation-norm-embedding}
\end{equation}
Since GNN performs message passing between neighboring nodes, splitting the graph into train and test subgraphs will prevent the test node information from leaking into the trained model. This is known as the inductive learning of GNN. Conversely, the entire graph can be kept intact in transductive learning, where train and test nodes participate in message passing and embedding learning. However, the test labels are not used during the supervised training process. While transductive learning can yield better test scores because of joint train and test node learning, inductive learning can provide more unbiased and generalizable outcomes due to the strict separation of train and test nodes. 

\subsection {Attention-based learning}

Edges in graphs are static and precomputed representations of connectivity between nodes. However, this connectivity may not represent the optimal relationship between nodes in the context of a learning objective.  An attention mechanism can dynamically learn an attention score between a pair of samples while optimizing the learning objective. Therefore, attention-based learning weeds out bad neighbors and focuses on the relevant parts of data to improve representation learning. Velikovic et al. introduce an attention mechanism in their graph attention model (GAT) that focuses on relevant neighboring nodes for better message passing~\citep{Velickovic2018}. Here, the attention mechanism dynamically updates the edge weights between relevant neighbors in each graph layer to emphasize their contribution to message passing. Instead of an adjacency matrix with static edge weights, the embedding for a single node is updated using dynamic attention scores, as shown in Equation~\ref{equation-gat}.
\begin{equation}
\Vec{h_p'}  = \sigma \left({\sum_{q \in \mathcal{N}_p} \alpha_{pq} W \Vec{h_q}} \right)
 \label{equation-gat}
\end{equation}
For a given node $p$, the set of all neighboring nodes is $\mathcal{N}_p$, and $h_q$ is the d-dimensional embedding of the neighboring node $q$. Here, $\alpha_{pq}$ is the attention score iteratively updated to measure the relevance between nodes $p$ and $q$. $W \in \Re^{d^{\prime} \times d}$ is the trainable weight matrix and $\sigma$ is the non-linear activation function. The attention-based GAT method uses unweighted or binary edge weights, whereas the GCN method uses weighted graphs. Similarly, GAT-based autoencoder (GATE)~\citep{Salehi2020} converts weighted graphs to unweighted graphs by replacing non-zero edge weights with 1 so that the model can learn attention scores from weak and strong edges regardless.

\subsection {Tabular data} 

Tabular data are structured in rows and columns of a matrix X $\in \Re^{n\times d}$ with $n$ samples and $d$ dimensional heterogeneous feature vector. Each sample ($X_i$) is represented in a feature vector form, $X_i = \{x_1, x_2, …, x_d\}$, where i = $\{1, 2,…,n\}$. Therefore,  tabular data are modeled using multivariate distributions $P(x_1, x_2, …, x_d)$ of heterogeneous variables. Unlike graph data, the samples in tabular data are assumed independent without any association or connection between rows. On the other hand, graph data may only have edge connection information between nodes without any feature vectors. However, tabular data sets often contain limited sample sizes and feature dimensionality, which may not take the full benefits of deep learning. Recent literature identifies heterogeneous tabular data as the last unconquered castle for deep learning~\citep{Kadra2021}, where traditional machine learning often performs better than deep learning methods~\citep{Perturbation, Kadra2021, Borisov2022}. Therefore, a robust deep representation of tabular data remains an open problem.
\subsection{Literature review}

In machine learning, samples in tabular data are assumed to be independent and identically distributed (i.i.d). The i.i.d assumption ignores any dependencies between samples. The identically distributed assumption considers similar data distribution to generalize the trained model on unseen test samples. The i.i.d assumption can be relaxed to learn useful associations between samples leveraging graph- and attention-based learning mechanisms. The relaxation of i.i.d assumption can be achieved by learning embedding from neighboring samples, which alters identical distribution assumption by aggregating neighborhood embeddings in representation learning~\citep{Hamilton2020}. We summarize relevant deep learning methods for tabular data below.

Because deep learning methods still fall short of traditional machine learning on Tabular data~\citep{Borisov2022,  ICACI}, recent studies propose new methods to improve the deep representation of tabular data. For example, attention-based methods can leverage information from other samples to improve embeddings for downstream classification tasks~\citep{Niu2021, Chaudhari2021}. A transformer-based attention model, feature tokenizer (FT) Transformer outperforms baseline methods on seven of eleven tabular data sets~\citep{Gorishniy2021}.  Non-Parametric Transformers, proposed for tabular data, outperform baseline models on eight classification data sets~\citep{Kossen2021}.  TabNet is an interpretable model that uses attention to select important features from tabular data~\citep{Arik2021} and outperforms other baseline methods on four out of five data sets and three out of six synthetic data sets.

In contrast, graph neural networks are not used beyond graph data sets. Several graph-based learning methods, including graph attention network (GAT)~\citep{Velickovic2018} and graph attention autoencoder (GATE)~\citep{Salehi2020}, have incorporated attention mechanisms to demonstrate superior performance over standard graph-based methods exclusively on graph data sets~\citep{Kipf2016}. Because of the i.i.d assumption, tabular data are rarely modeled as graph data to investigate the effectiveness of graph-based learning. Few studies have proposed graph-based models for tabular data, including bipartite graphs (GRAPE) \citep{You2020} and Propagation to Enhance Tabular data prediction (PET) \citep{Du2022}, and multiplex graph (TabGNN) \citep{Guo2021}. In the GRAPE approach, tabular data ($X_{m \times n}$) is modeled as a bipartite graph $G = (V_s, V_f, E)$ where $V_s$ and $V_f$ denote sets of sample and feature nodes and E represents all undirected edges connecting sample and feature nodes. An edge (i, j) weight represents the j-th feature value for the i-th sample. In contrast, the PET method uses unweighted edges with feature nodes representing the feature values. However, these approaches do not model the between-sample relationships. Lastly, TabGNN connects sample nodes that share the same values for the categorical variables. Here, multiple categorical features result in multiple types of edges. However, this method leaves out numerical variables in modeling graph structures of tabular data. To the best of our knowledge, graph neural networks have not been considered in learning between sample relationships in tabular data with numerical variables or compared against traditional machine learning or attention-based learning methods.






\section {Methods}

This section provides methods to create the adjacency matrix for graph based learning of tabular data. The section summarizes the baseline and evaluation methods for comparing several deep learning alternatives with superior machine learning methods on tabular data. 

\subsection {Modeling tabular data in graphs} \label{section_table2graph}

A graph data set includes edge weights and adjacency matrix for graph-based learning. Therefore, an adjacency matrix is formed using tabular data samples to train GNN. There is no single or best approach to forming an adjacency matrix with edge weights. The adjacency matrix for tabular data is first computed by measuring 1) Euclidean distances and 2) cosine similarity between each pair of sample feature vectors. The distance or similarity metric scale is normalized column-wise using minimum and maximum values. The scaled Euclidean distance values are transformed into similarity by taking their absolute difference from unity.

Finally, a threshold is used on the similarity value to drop weak and spurious connections in the graph. For the unweighted or binary graph scenario, a similarity score above or equal to the threshold will be replaced by one; otherwise by zero. In contrast, all similarity values are preserved in the adjacency matrix for a weighted graph after pruning the below-threshold values to zero. We use inductive graph learning to ensure the separation between train and test nodes during supervised classification.

\subsection{Baseline methods}
We compare tabular data classification performances of nine baseline methods: three traditional machine learning models, three recently proposed attention-based deep methods for tabular data, and three graph-based neural networks. Machine learning methods include logistic regression (LR), gradient boosting tree (GBT), and multi-layer perceptron (MLP). The attention-based tabular data learning methods include FT-Transformer (FTT) \citep{Gorishniy2021}, TabNet \citep{Arik2021}, and Non-Parametric Transform (NPT) \citep{Kossen2021}. These baseline methods are compared against three graph neural network methods: Graph Convolution Network (GCN) \citep{Kipf2016}, Graph Attention Network (GAT) \citep{Velickovic2018}, and Graph Attention Autoencoder (GATE) \citep{Salehi2020}. 

\subsubsection{TabNet} TabNet is motivated by ensemble decision trees to create sequential blocks termed as decision steps~\citep{Arik2021}. Each block uses an attention mechanism to select a subset of features for learning the embedding and passes the feature subset to the next block. The next block selects a subset of the feature subset and creates another embedding. All embeddings from the decision steps are aggregated in the final embedding for downstream classification. In each decision step, the sparsity of feature selection is calculated in the form of entropy. The model is trained until the sum of prediction loss and sparse entropy converges. 

\subsubsection{FT-Transformer}  Feature Tokenizer + Transformer (FT-Transformer) is a deep tabular model that uses multiheaded self-attention (MHSA) (attention between features) to refine feature embeddings as tokenized features~\citep{Gorishniy2021}. A linear layer maps the final embedding to target classification labels. The model is trained in a supervised manner until the convergence of the classification loss. 

\subsubsection{NPT} Non-parametric transformer (NPT) predicts target labels by learning to attend between samples and features~\citep{Kossen2021}. This differs from traditional models, where the target label is a function of the feature vector alone. Instead, the NPT model predicts missing entries masked stochastically in the training data. NPT uses stacked MHSA layers to attend between sample and feature representations. The model is trained until the sum of missing value reconstruction loss and target loss  converges. 

\subsubsection{GCN} Graph convolutional network (GCN) is built using a sequence of graph layers \citep{Kipf2016} that perform the message passing mechanism, as discussed in Section~\ref{background}. However, instead of a normalized adjacency matrix, it uses a normalized Laplacian matrix, also known as symmetric neighborhood normalization. Graph convolution networks differ from convolutional neural networks that work on image data. While CNN layers aggregate pixel information from neighbors arranged in a regular grid, graph convolution layers aggregate irregular neighboring node features. 

\subsubsection{GAT} Graph Attention Network (GAT) is based on GCN model but learns attention scores between node pairs using features~\citep{Velickovic2018}. The attention scores are used instead of edge weights to aggregate the neighborhood embeddings as individual node embedding. The authors draw motivation from MHSA and use multi-head to compute k parallel attention scores between connected nodes. The final layer maps the final embedding to classification targets. The model is trained until the supervised loss converges. The authors report superior performance of their GAT method to the GCN method in both transductive and inductive learning on three graph data sets.

\subsubsection{GATE} Graph Attention Autoencoder (GATE) is an autoencoder model built using GAT graph layers in its encoder and decoder stacks~\citep{Salehi2020}. An unweighted (binary) adjacency matrix is used to initiate the computing of attention scores. The model is trained to minimize the node feature reconstruction loss and maximize the similarity between neighboring node representations. The embedding from the last encoder layer is used for downstream classification tasks. The authors report superior performance of the GATE method compared to the GAT method on the same three benchmark graph data sets.

\subsection {Data sets}
A summary of the ten tabular data sets used in this paper is provided in Table~\ref{table_dataset}.  These data sets are representative of distinct application domains. Except for the Olive dataset\footnote{https://search.r-project.org/CRAN/refmans/FlexDir/html/oliveoil.html}, all other data sets are sourced from the UCI machine learning repository. Only numeric features are considered for this study. Except for the satellite and malware data sets, others have less than one thousand samples. The data dimensionality ranges from five to 753. Only three data sets: Parkinson's, breast cancer, and malware, have binary classification labels.

\begin{table}[]
\scalebox{0.9}{
\begin{tabular}{lcccc}
\toprule
Data set & Samples & Features & Classes & F-S ratio \\ \midrule
Wine & 178 & 13 & 3 & 0.073 \\
Olive & 572 & 10 & 3 & 0.017 \\
Vehicle & 846 & 18 & 4 & 0.021\\
Satellite & 6435 & 36 & 6 & 0.006\\
Parkinson & 756 & 753 & 2 & 0.996\\
Breast cancer & 569 & 30 & 2 & 0.053 \\
Dermatology & 358 & 34 & 6 & 0.095\\
Mice & 552 & 77 & 8 &0.140\\
Malware & 4464 & 241 & 2 &0.054\\
Synthetic & 200 & 5 & 4 &0.025\\
\bottomrule
\end{tabular}
\caption{Summary of tabular data sets used to evaluate graph and attention-based learning methods in this paper. F-S ratio = feature-sample ratio.}
\label{table_dataset}
}
\end{table}

\subsection{Model evaluation}
The classification performance is reported using the average F1 weighted score in a 10-fold cross-validation scheme. In 10-fold cross-validation, a test fold (10\%) is set apart, and the remaining seven and two folds are used for training and validation, respectively. A predefined random state seed for data split and cross-validation reproduces the same train and test samples for all experiments. We use the default hyperparameter setting for attention and graph-based methods. However, hyperparameters are tuned for the traditional machine learning methods. For graph-based learning, we tune the construction of the adjacency matrix by varying the cut-off value between  0 and 1 at 0.1 intervals to determine the weak connections (See Section \ref{section_table2graph}) and the selection between weighted and unweighted graph.

\newcolumntype{Y}{p{1.1cm}}
\begin{table*}[t]
\scalebox{0.8}{
\begin{tabular}{lYYYYYYYYYY}
\toprule
 & Breast cancer & Derma-tology & Malware & Mice & Olive & Parkin-son's & Satellite & Synthetic & Vehicle & Wine \\ \midrule
LR & 0.970 $(0.02)^{}$ & 0.966 $(0.03)^{}$ & 0.985 $(0.00)^{}$ & 0.998 $(0.01)^{}$ & 0.854 $(0.04)^{}$ & 0.830 $(0.03)^{}$ & 0.828 $(0.01)^{}$ & 0.970 $(0.03)^{}$ & 0.789 $(0.03)^{}$ & 0.972 $(0.03)^{}$ \\ \midrule
GBT & 0.949 $(0.03)^{}$ & \bfseries 0.969 $(0.02)^{}$ & 0.991 $(0.00)^{}$ & 0.961 $(0.03)^{}$ & 0.868 $(0.05)^{}$ & 0.848 $(0.07)^{}$ & 0.909 $(0.01)^{}$ & 0.980 $(0.02)^{}$ & 0.734 $(0.02)^{}$ & 0.937 $(0.07)^{}$ \\ \midrule
MLP & 0.972 $(0.02)^{}$ & 0.557 $(0.11)^{}$ & \bfseries 0.993 $(0.00)^{}$ & 0.925 $(0.09)^{}$ & \bfseries 0.879 $(0.05)^{}$ & 0.846 $(0.05)^{}$ & 0.736 $(0.06)^{}$ & 0.659 $(0.01)^{}$ & 0.746 $(0.14)^{}$ & \bfseries 0.983 $(0.03)^{}$ \\ \midrule
FTT & 0.970 $(0.02)^{}$ & 0.960 $(0.03)^{\dagger}$ & 0.987 $(0.00)^{\dagger}$ & \bfseries 1.000 $(0.00)^{\diamond\dagger}$ & 0.844 $(0.06)^{\dagger}$ & 0.838 $(0.07)^{}$ & 0.911 $(0.01)^{\star\dagger}$ & \bfseries 1.000 $(0.00)^{\star\diamond\dagger}$ & \bfseries 0.811 $(0.03)^{\diamond}$ & 0.978 $(0.04)^{}$ \\ \midrule
TabNet & 0.942 $(0.02)^{\star\dagger}$ & 0.941 $(0.03)^{\diamond\dagger}$ & 0.978 $(0.01)^{\star\diamond\dagger}$ & 0.876 $(0.04)^{\star\diamond}$ & 0.845 $(0.06)^{\dagger}$ & 0.818 $(0.05)^{}$ & 0.879 $(0.01)^{\star\diamond\dagger}$ & 0.980 $(0.03)^{\dagger}$ & 0.698 $(0.04)^{\star\diamond}$ & 0.931 $(0.03)^{\star\dagger}$ \\ \midrule
NPT & \bfseries 0.972 $(0.02)^{\diamond}$ & 0.959 $(0.02)^{\dagger}$ & - & - & 0.858 $(0.06)^{}$ & - & \bfseries 0.915 $(0.01)^{\star\diamond\dagger}$ & \bfseries 1.000 $(0.00)^{\star\diamond\dagger}$ & 0.793 $(0.05)^{\diamond}$ & 0.955 $(0.04)^{}$ \\ \midrule
GATE\_CB & 0.968 $(0.02)^{}$ & 0.847 $(0.06)^{\star\diamond\dagger}$ & 0.987 $(0.01)^{\dagger}$ & \bfseries 1.000 $(0.00)^{\diamond\dagger}$ & 0.816 $(0.05)^{\star\diamond\dagger}$ & 0.793 $(0.04)^{\star\diamond\dagger}$ & 0.784 $(0.01)^{\star\diamond}$ & 0.870 $(0.16)^{\diamond\dagger}$ & 0.639 $(0.06)^{\star\diamond}$ & 0.954 $(0.04)^{}$ \\ \midrule
GATE\_EB & 0.956 $(0.02)^{}$ & 0.877 $(0.06)^{\star\diamond\dagger}$ & 0.985 $(0.01)^{\dagger}$ & 0.497 $(0.06)^{\star\diamond\dagger}$ & 0.622 $(0.14)^{\star\diamond\dagger}$ & 0.787 $(0.04)^{\star\diamond\dagger}$ & 0.783 $(0.01)^{\star\diamond}$ & 0.970 $(0.06)^{\dagger}$ & 0.740 $(0.05)^{\star}$ & 0.977 $(0.03)^{}$ \\ \midrule
GAT\_CB & 0.964 $(0.02)^{}$ & \bfseries 0.969 $(0.04)^{\dagger}$ & 0.983 $(0.01)^{\diamond\dagger}$ & 0.972 $(0.03)^{\star}$ & 0.817 $(0.04)^{\star\diamond\dagger}$ & 0.857 $(0.04)^{\star}$ & 0.642 $(0.16)^{\star\diamond}$ & 0.978 $(0.05)^{\dagger}$ & 0.594 $(0.06)^{\star\diamond\dagger}$ & 0.965 $(0.05)^{}$ \\ \midrule
GAT\_EB & 0.958 $(0.03)^{}$ & 0.962 $(0.03)^{\dagger}$ & 0.982 $(0.00)^{\diamond\dagger}$ & 0.985 $(0.02)^{}$ & 0.820 $(0.04)^{\star\diamond\dagger}$ & 0.844 $(0.05)^{}$ & 0.841 $(0.01)^{\star\diamond\dagger}$ & 0.995 $(0.02)^{\dagger}$ & 0.608 $(0.05)^{\star\diamond\dagger}$ & 0.961 $(0.06)^{}$ \\ \midrule
GCN\_C & 0.956 $(0.03)^{}$ & 0.960 $(0.03)^{\dagger}$ & 0.985 $(0.01)^{\dagger}$ & 0.887 $(0.06)^{\star\diamond}$ & 0.725 $(0.09)^{\star\diamond\dagger}$ & \bfseries 0.860 $(0.04)^{\star}$ & 0.717 $(0.03)^{\star\diamond}$ & 0.959 $(0.11)^{\dagger}$ & 0.543 $(0.05)^{\star\diamond\dagger}$ & 0.931 $(0.10)^{}$ \\ \midrule
GCN\_CB & 0.959 $(0.03)^{}$ & 0.951 $(0.03)^{\dagger}$ & 0.985 $(0.01)^{\dagger}$ & 0.881 $(0.06)^{\star\diamond}$ & 0.738 $(0.06)^{\star\diamond\dagger}$ & 0.845 $(0.03)^{}$ & 0.684 $(0.13)^{\star\diamond}$ & 0.957 $(0.10)^{\dagger}$ & 0.538 $(0.05)^{\star\diamond\dagger}$ & 0.960 $(0.06)^{}$ \\ \midrule
GCN\_E & 0.954 $(0.03)^{}$ & 0.953 $(0.03)^{\dagger}$ & 0.979 $(0.01)^{\star\diamond\dagger}$ & 0.904 $(0.05)^{\star\diamond}$ & 0.746 $(0.08)^{\star\diamond\dagger}$ & 0.852 $(0.03)^{}$ & 0.782 $(0.05)^{\star\diamond\dagger}$ & 0.990 $(0.02)^{\dagger}$ & 0.589 $(0.06)^{\star\diamond\dagger}$ & 0.943 $(0.07)^{\dagger}$ \\ \midrule
GCN\_EB & 0.970 $(0.02)^{}$ & 0.936 $(0.04)^{\star\diamond\dagger}$ & 0.980 $(0.01)^{\diamond\dagger}$ & 0.897 $(0.06)^{\star\diamond}$ & 0.787 $(0.07)^{\star\diamond\dagger}$ & 0.835 $(0.05)^{}$ & 0.808 $(0.03)^{\diamond\dagger}$ & 0.975 $(0.03)^{\dagger}$ & 0.613 $(0.05)^{\star\diamond\dagger}$ & 0.966 $(0.08)^{}$ \\ \bottomrule
\end{tabular}
\caption{Average F1 scores. For graph-based models, the adjacency matrix is constructed with C = cosine or E= Euclidean similarity. B denotes an unweighted adjacency matrix. Otherwise, adjacency matrices are weighted. $\star$, $\diamond$, and $\dagger$ indicate that the average F1 score is significantly different ($p<0.05$) compared to LR, GBT, and MLP, respectively.}
\label{table_performance}
}
\end{table*}

Our MLP with three hidden layers (256-128-64) is trained with a learning rate of 0.001 and a batch size of 64. For GBT, the hyperparameters are the number of estimators (50, 80, or 110) and maximum tree depth (2, 5, 10, 15). The LR classifier is tuned for L1 and L2 regularization and the inverse of the regularization strength parameter (0.01, 0.05, 0.5, 0.8, 1, 5). However, we train on the nine data folds for unsupervised learning (e.g., the GATE method) and report the classification accuracy on the left-out test fold. All neural network-based models are trained for 1000 epochs with patience 100. The validation and training losses are used for early stopping in supervised and unsupervised training, respectively. Scores from individual test folds are used to perform a Wilcoxon-ranked test to compare the performance of the methods statistically.

\section {Results}
The performance of each model on different tabular data sets is presented in Table \ref{table_performance}. The statistical significance in classification performance ($p\leq0.05$) compared to traditional models (LR, GBT, and MLP) is also indicated in the Table. 

\subsection {Performance of Graph Neural Networks}
GATE and GAT methods with unweighted cosine similarity metrics perform better than GBT on the breast cancer data set. The same GAT method ties with GBT as the best classifier on the dermatology data set. GATE with an unweighted cosine metric is on par with the LR classifier for the malware data set. The same GATE method is statistically superior to GBT and MLP as the best-performing method on the mice data set. However, graph-based learning is statistically inferior to traditional machine learning methods (GBT, LR, and MLP) on the olive data set. The GAT method is slightly superior to GBT and LR methods on the Parkinson's data set, although the improvement is not statistically significant. Overall, the GCN method with weighted cosine similarity outperforms all other methods on the Parkinson's data set. However, graph-based methods are statistically inferior to machine learning methods on the satellite data set. The GAT method is on par with traditional classifier models (GBT, LR) for the synthetic data set but statistically superior to MLP. The GATE method with an unweighted Euclidean distance metric is better than LR and GBT. 


\begin{table*}[]
\scalebox{0.8}{
\begin{tabular}{lcccccccccc}
\toprule
 & Breast cancer & Dermatology & Malware & Mice & Olive & Parkinson's & Satellite & Synthetic & Vehicle & Wine \\ \midrule 
LR & 3 & 3 & 5 & 3 & 4 & 6 & 6 & 7 & 3 & 4 \\ \midrule
GBT & 8 & 1 & 2 & 5 & 2 & 3 & 3 & 5 & 6 & 8 \\ \midrule
MLP & 1 & 9 & 1 & 6 & 1 & 4 & 9 & 9 & 4 & 1 \\ \midrule
FTT & 3 & 4 & 3 & 1 & 6 & 5 & 2 & 1 & 1 & 2 \\ \midrule
TabNet & 9 & 7 & 8 & 8 & 5 & 7 & 4 & 5 & 7 & 9 \\ \midrule
NPT & 1 & 6 & - & - & 3 & - & 1 & 1 & 2 & 7 \\ \midrule
GATE & 6 & 8 & 3 & 1 & 8 & 8 & 8 & 7 & 5 & 3 \\ \midrule
GAT & 7 & 1 & 7 & 4 & 7 & 2 & 5 & 3 & 9 & 6 \\ \midrule
GCN & 3 & 4 & 5 & 7 & 9 & 1 & 7 & 4 & 8 & 5 \\ \bottomrule
\end{tabular}
\caption{Rank ordering of traditional machine learning (LR, GBT, MLP), deep learning (FTT, TabNet, NPT), and graph neural networks (GATE, GAT, GCN) in the classification performance across ten tabular data sets. In case of ties, the next rank is skipped.}
\label{table_rank}
}
\end{table*}

\subsection{Performance of deep attention models}
TabNet is a deep learning method specifically proposed for tabular data. However, the GATE method outperforms the TabNet method on five of ten data sets. Notably,  The GAT method outperforms TabNet on seven of ten data sets. The NPT method performs the best on the breast cancer, satellite, and synthetic data sets. However, it fails to converge on the malware, mice, and Parkinson's data sets. These are the three data sets with the highest data dimensionality. Yet the NPT method outperforms GAT on five of ten data sets. Overall, the transformer-based FTT method is superior to graph-based methods, with some exceptions. For instance, FTT is on par with graph-based learning on the breast cancer, malware, mice, and wine data sets but inferior on the dermatology and Parkinson's data sets.

\subsection {Best performing methods}

Table~\ref{table_rank} summarizes the rank ordering of different learning methods for the tabular data sets. The LR classifier does not perform the best on any data sets but challenges other deep and attention-based learning on several data sets. GBT appears as one of the methods for the dermatology data set. MLP is the best method on three of the ten data sets: malware, olive, and wine. Among the deep tabular methods, FTT achieves the best F1 scores on three of ten data sets: mice, synthetic, and vehicle. TabNet does not achieve the best F1 score on any data set. NPT achieves the best F1 scores on the breast cancer (tied with MLP), satellite, and synthetic (tied with FTT) data sets. GAT with unweighted cosine similarity is the best on the dermatology data set. GATE with unweighted cosine similarity is the best method for the mice data set. Finally, the GCN with weighted cosine similarity metric performs the best on the Parkinson's data set.

\section {Discussion of results}

This paper investigates the effectiveness of learning tabular data representation by leveraging between-sample relationships, which is generally ignored due to the i.i.d assumption in machine learning. Traditional machine learning methods are compared against attention and graph networks modeling the between-sample relationship. The findings of the paper are summarized as follows. First, no single method performs the best on all tabular data sets. This confirms the knowledge that the performance of a method depends on the data or domain problem. Second, graph neural networks, such as GAT, GATE, and GCN methods with cosine similarity metrics, have shown promising results, outperforming state-of-the-art deep tabular data learning methods on several data sets. This reveals that some domain data sets may have useful relationships between samples defying the i.i.d assumption. Third, attention between samples (NPT method) and between-sample relationships in GNN methods secure the best ranks on six of the ten tabular data sets. This observation suggests that the independent sample assumption should be relaxed to improve tabular data representation. 

The deep learning methods proposed for tabular data (FTT, NPT, TabNet) are among the best methods for the data sets with the largest sample sizes (vehicle, satellite, and malware). Notably, unlike our experiments, these methods are evaluated by selectively using data sets with a large sample size. For example, the TabNet article selectively uses tabular data sets with sample sizes ranging from 41 thousand to 11 million. Therefore, the performance of similar methods may be limited to data sets with large sample sizes. Furthermore, the deep tabular methods create larger dimensional embeddings than the input dimension. However, our experiments find that NPT is very memory expensive and unstable on data sets with high dimensional features (mice (77), malware (241), and Parkinson's (753)). Notably, the data dimensionality in Tabnet, FTT, and NPT range from 11 to 54, 8 to 2000 , and 10 to 54, respectively. This explains that FTT is effective for a wide range of data dimensionality, which our results support. 

One interesting trend in the performance of GNN methods may be explained using the number of features to the number of samples ratio (F-S ratio) for each data set, as presented in Table~\ref{table_dataset}. For the data sets with the highest ratios (Parkinson's: 0.996, Dermatology: 0.095, and Mice: 0.14), one of the GNN methods ranks as the best. In contrast, GNN methods perform poorly on data sets with the lowest ratios (Olive: 0.017, Vehicle: 0.02, Satellite: 0.006). Therefore, GNN methods can be chosen for tabular data sets with limited samples but large dimensionality. A general trend suggests that a binary adjacency matrix built from cosine similarity is superior to other methods in learning tabular data with GNN.

\section {Conclusions}
This paper conducts one of the first studies to investigate the effectiveness of learning between-sample relationships in tabular data conventionally considered independent. Graph and attention-based models are used to learn between-sample relationships for downstream classification tasks, unlike traditional machine learning methods. Our findings suggest that between-sample relationships via graph edges or attention modeling can outperform traditional machine-learning methods on seven of the ten data sets. This finding is important because it shows an alternative approach to improving deep learning of tabular data sets when traditional machine learning outperforms deep learning methods.  


\section*{Acknowledgments}

Research reported in this publication was supported in part by the National Library of Medicine of the National Institutes of Health under Award Number R15LM013569 and was partially sponsored by the Air Force Office of Scientific Research under Grant Number W911NF-23-1-0170. The content is solely the responsibility of the authors and does not necessarily represent the official views of the National Institutes of Health and should not be interpreted as representing the official policies, either expressed or implied, of
the Army Research Office or the U.S. Government.

\bibliographystyle{unsrt}

\bibliography{myBib}

\end{document}